\documentclass[10pt,twocolumn,letterpaper]{article}

\pdfoutput=1

\usepackage{cvpr}
\usepackage{times}
\usepackage{graphicx}
\usepackage{amsmath,amssymb}
\usepackage[tight]{subfigure}
\usepackage{hyperref}
\usepackage{algorithmic}
\usepackage{algorithm}
\usepackage{balance}

\hypersetup{%
  pdftitle={Efficient Information Theoretic Clustering on Discrete Lattices},
  pdfauthor={Christian Bauckhage and Kristian Kersting},
  pdfsubject={pattern recognition, image analysis, clustering, data science, statistics},
  pdfkeywords={information theoretic clustering, convolution},
  colorlinks=true,
  bookmarks=false}

\graphicspath{./}

\newcommand{\mat}[1]{\boldsymbol{\mathbf{#1}}}

\cvprfinalcopy

\title{Efficient Information Theoretic Clustering on Discrete Lattices}
\author{Christian Bauckhage$^{1,3}$ and Kristian Kersting$^{2,3}$\\
\emph{$^{1}$B-IT, University of Bonn, 53113 Bonn , Germany}\\
\emph{$^{2}$IGG, University of Bonn, 53115 Bonn , Germany}\\
\emph{$^{3}$Fraunhofer IAIS, 53754 St. Augustin, Germany}}

\begin{document}
\maketitle

\begin{abstract}
We consider the problem of clustering data that reside on discrete, low dimensional lattices. Canonical examples for this setting are found in image segmentation and key point extraction. Our solution is based on a recent approach to information theoretic clustering where clusters result from an iterative procedure that minimizes a divergence measure. We replace costly processing steps in the original algorithm by means of convolutions. These  allow for highly efficient implementations and thus significantly reduce runtime. This paper therefore bridges a gap between machine learning and signal processing.
\end{abstract}

\section{Motivation and Background}

Clustering is an important tool of the trade in data mining and pattern recognition. Numerous algorithms have been proposed and an extended review of their characteristics and merits would be beyond the scope of this paper. However, we note that many algorithms used to cluster numerical data tacitly assume the data to be drawn from distributions that are continuous.

Alas, many data are explicitly discrete. Consider, for instance, the pixels of a standard digital image. Owing to modern recording technologies, domain and range of such images are discrete and we may distinguish three elementary cases:

(i) In a binary (black and white) image, each pixel is switched either on or off. Pixels that are switched off need not be represented; pixels that are switched on are characterized by their two-dimensional grid coordinate $[u, v]^T$.

(ii) In an intensity (gray value) image, pixels are located on a two-dimensional grid and assume discrete gray values; they are thus fully characterized by a three-dimensional grid coordinate $[u, v, g]^T$.

(iii) In a color (RGB) image, each pixel has a fixed location on a two-dimensional grid and maps to a discrete three-dimensional color vector; color pixels are thus fully characterized by a five-dimensional grid coordinate $[u, v, r, g, b]^T$.

Digital image therefore correspond to sets of data points that reside on low dimensional lattices. If such data need to be clustered into coherent groups, it is common practice to resort to algorithms that perform continuous optimization and thus will likely produce cluster centers that are located in between lattice points. In most practical applications, this is perfectly acceptable and our concern in this paper is therefore not with precision but with efficiency.

In this paper, we adapt an algorithm for continuous, information theoretic clustering to data on discrete lattices. Our basic idea is to connect recent machine learning techniques with well established methods for digital signal processing.

We show that recasting essential steps of information theoretic clustering in terms of efficient convolution operations significantly reduces its runtime and enables us to quickly compute cluster centers in image data. To simplify our discussion, we focus on applications in binary image processing and provide qualitative examples to illustrate behavior and performance of the algorithm. Quantitative experiments on a large data base of binary images reveal that our accelerated algorithm is two orders of magnitude faster than the original one. Moreover, we demonstrate that weighted clustering is straightforward in our framework.

Next, we review information theoretic clustering and discuss its properties. Then, we derive our accelerated version and present experimental results that underline its favorable performance. Finally, we discuss how the proposed algorithm immediately allows for efficient weighted clustering and conclude this paper with a summary and an outlook to practical applications of our approach.

\section{Information Theoretic Clustering}

The particular approach to information theoretic clustering (ITC) we consider in this paper was introduced in \cite{LehnSchioler2005-VQU,Rao2009-MSA,Rao2007-ITV}. Given a set
\begin{equation*}
\mathcal{X} = \{x_1, \dots, x_N\}, \quad x_i \in \mathbb{R}^d
\end{equation*}
of data  points drawn from a continuous domain, it applies an entropy-based measure to determine $M \ll N$ clusters in the data. Similar to $k$-means clustering, ITC starts from a random initialization of the cluster centers
\begin{equation*}
\mathcal{W} = \{w_1, \ldots, w_M\}, \quad w_j \in \mathbb{R}^d
\end{equation*}
and iteratively updates these \emph{codebook vectors} in order to minimize the entropy measure under consideration.

\begin{figure*}[t!]
    \begin{center}
    \subfigure[a set $\mathcal{X}$ of 2D data points]{%
        \includegraphics[width=0.24\textwidth]{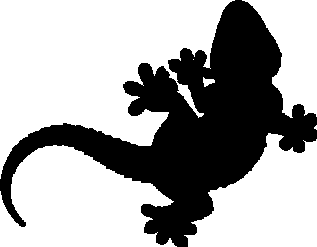}} \hfill
    \subfigure[pdf $p(x)$ of $\mathcal{X}$]{%
        \includegraphics[width=0.24\textwidth]{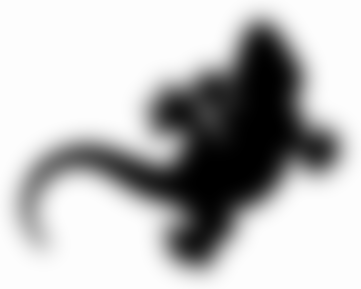}} \hfill
    \subfigure[initial pdf $q(x)$ of $\mathcal{W}$]{%
        \includegraphics[width=0.24\textwidth]{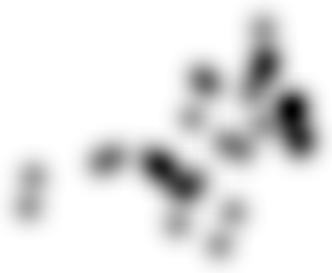}} \hfill
    \subfigure[optimized pdf $q(x)$ of $\mathcal{W}$]{%
        \includegraphics[width=0.24\textwidth]{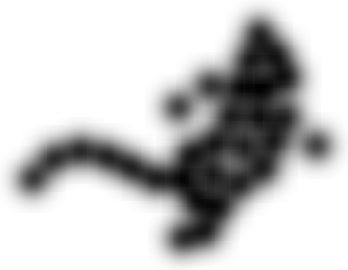}}
    \end{center}
    \caption{\label{fig:densities} Information theoretic clustering of a binary image: (a) a binary shape can be understood as a set $\mathcal{X} = \{x_i\}_{i=1}^N$ of 2D data points; (b) visualization of a Parzen estimate of the probability density function $p(x)$ of $\mathcal{X}$; (c) a randomly initialized set $\mathcal{W}$ of codebook vectors (30 in this case) typically has a pdf $q(x)$ that diverges from $p(x)$; (d) an iterative minimization the Cauchy-Schwartz divergence of $p(x)$ and $q(x)$ moves the codebook vectors towards modes of $p(x)$.}
\end{figure*}

\subsection{Iterative Entropy Minimization}
ITC assumes that two probability density functions $p(x)$ and $q(x)$ are available that characterize the sets $\mathcal{X}$ and $\mathcal{W}$, respectively. Given these densities, ITC optimizes the placement of codebook vectors by means of minimizing the Cauchy-Schwartz divergence
\begin{equation}
D_{cs}(\mathcal{X}, \mathcal{W}) = 2H(\mathcal{X}; \mathcal{W}) - H(\mathcal{W}) - H(\mathcal{X})
\label{eq:cauchyschwartz}
\end{equation}
between $p(x)$ and $q(x)$. Note that the last term of $D_{cs}$ does not depend on $\mathcal{W}$. The entropies in its first and second term are given by
\begin{align*}
H(\mathcal{X}; \mathcal{W})
    & = -\log \int p(x) \, q(x) \, dx  \;
      \stackrel{!}{=} -\log V(\mathcal{X}; \mathcal{W})
\intertext{and}
H(\mathcal{W})
    & = -\log \int q^2(x) \, dx \;
      \stackrel{!}{=} -\log  V(\mathcal{W})
\end{align*}
and are known as Renyi's cross entropy between $p(x)$ and $q(x)$ and Renyi's entropy of $q(x)$, respectively \cite{Renyi1961-OMO}.

The fundamental idea in \cite{LehnSchioler2005-VQU,Rao2009-MSA,Rao2007-ITV} is to model the densities $p(x)$ and $q(x)$ using their Parzen estimates
\begin{align}
p(x) & = \frac{1}{N} \sum_{i = 1}^N G_{\xi}(x - x_i)    \label{eq:dens1}\\
q(x) & = \frac{1}{M} \sum_{j = 1}^M G_{\omega}(x - w_j) \label{eq:dens2}
\end{align}
where $G_{\xi}(x)$ and $G_{\omega}(x)$ are Gaussian kernels of variance $\xi^2$ and $\omega^2$, respectively.

Applying the Gaussian product theorem and using properties of the so called overlap integral, $V(\mathcal{X}; \mathcal{W})$ and $V(\mathcal{W})$ can be written as
\begin{align*}
    \int p(x) q(x) \, dx
        & = \frac{\sum_{i,j} \! \int \! G_{\xi}(x \! - \! x_i) \, G_{\omega}(x \! - \! w_j) \, dx}{MN} \\
        & = \frac{1}{MN}\sum_{i = 1}^N \sum_{j = 1}^M  G_{\tau}(x_i - w_j)\\
    \intertext{and}
    \int q^2(x) \, dx
        & = \frac{\sum_{i,j} \! \int \! G_{\omega}(x \! - \! w_i) \, G_{\omega}(x \! - \! w_j) \, dx}{M^2} \notag \\
        & = \frac{1}{M^2}\sum_{i = 1}^M  \sum_{j = 1}^M G_{\rho}(w_i - w_j)
\end{align*}
where $\tau^2 = \xi^2 +  \omega^2$ and $\rho^2 = 2 \omega^2$.

Therefore, differentiating $D_{cs}(\mathcal{X}, \mathcal{W}) $ with respect to $w_k$, equating to zero, and rearranging the resulting terms yields the following fix point update rule for every codebook vector in $\mathcal{W}$
\begin{align}
    w_k^\text{new} =
    & \frac{\sum_{j=1}^N G_{\tau} (x_j - w_k) \, x_j}{\sum_{j=1}^N G_{\tau}(x_j - w_k)} \nonumber \\
    & - c \cdot \frac{\sum_{j=1}^M G_{\rho} (w_j - w_k) \, w_j}{\sum_{j=1}^N G_{\tau}(x_j - w_k)} \nonumber \\
    & + c \cdot \frac{\sum_{j=1}^M G_{\rho} (w_j - w_k)}{\sum_{j=1}^N G_{\tau}(x_j - w_k)} \, w_k \label{eq:origupdate}
\end{align}
where the constant $c = \frac{N}{M} \frac{V(\mathcal{X}; \mathcal{W})}{V(\mathcal{W})}$.

\subsection{Properties of ITC}
In \cite{Rao2009-MSA}, it is shown that the popular mean shift procedure \cite{Cheng1995-MSM,Comaniciu2002-MSA} is a special case of ITC. The cross entropy term $\log V(\mathcal{X}; \mathcal{W})$ in the Cauchy-Schwartz divergence causes ITC to be a mode seeking algorithm. The quantity $\log V(\mathcal{W})$, however, can be understood as the potential energy of a repellent force between the codebook vectors \cite{Rao2009-MSA}. After convergence of the algorithm, the codebook vectors will therefore be located close to local modes but because of the repellent force they will not have collapsed into only a few modes.

Figure~\ref{fig:densities} illustrates how ITC determines a codebook of 30 vectors from a binary shape image. Every foreground pixel of the original image in the leftmost panel is understood as a 2D vector $x_i \in \mathbb{R}^2$; together they form the data set $\mathcal{X}$. Using (\ref{eq:dens1}) to compute an estimate of the density $p(x)$ of $\mathcal{X}$ yields the result shown in the second panel. The third panel demonstrates that a random initialization of a set of codebook vectors $\mathcal{W}$ typically yields a density function $q(x)$ that diverges from the density $p(x)$ of the data points. Iterative updates of the codebook vectors using (\ref{eq:origupdate}) will minimize this divergence. The rightmost panel of the figure shows how, upon termination of the ITC algorithm, the elements in $\mathcal{W}$ have moved towards local modes of $\mathcal{X}$ and that the two densities $p(x)$ and $q(x)$ are now in close agreement. Since the approach prevents codebook vectors from collapsing into a few modes, the final distribution $q(x)$ is found to trace out local principal curves of the binary shape.

For $N$ data points and $M$ codebook vectors, every iteration of ITC requires efforts of $O(MN)$. However, in contrast to related clustering algorithms of similar complexity such, for instance, $k$-means, there is a large constant that factors into the overall runtime of ITC. Each of the $k = 1, \ldots, M$ updates in (\ref{eq:origupdate}) requires the computation of $N$ distances $\Vert x_j - w_k \Vert^2$ and $M$ distances $\Vert w_j - w_k \Vert^2$ and each distance computation entails the evaluation of an exponential function in a Gaussian kernel.

With respect to modern practical applications, for instance in large scale image processing, the algorithm will therefore be too slow to be useful. It would neither allow for an efficient processing of a few very large images nor would it apply to the processing of large amounts of simple binary images of only moderate size.

\section{ITC on Discrete Lattices}

Addressing the efficiency concerns raised in the previous section, we now derive an alternative update rule for information theoretic clustering. Our key contribution at this point is to model the Parzen density estimates in (\ref{eq:dens1}) and (\ref{eq:dens2}) using methods from signal processing. Mathematically, our result is equivalent to the original formulation but it is cast in a form that allows for avoiding repeated computations of distances and exponentials. Especially for data that reside on discrete lattices our new formulation of ITC enables highly efficient implementations based on discrete filter masks and therefore considerably accelerate the procedure. To ease our discussion, we will again focus on examples from binary image processing but the methods discussed below immediately generalize to higher dimensional lattices, too.

\subsection{Entropy Minimization Using Convolutions}
Recall that the convolution of two functions $f(x)$ and $g(x)$ is defined as
\begin{equation*}
(f \ast g) (x) = \int f(y) \, g(x-y) \, dy
\end{equation*}
where integration is over the whole, possibly higher dimensional domain.

To begin our derivation, we note that the convolution of a function $f$ with a shifted delta impulse $\delta(x-\mu)$ shifts the function
\begin{equation*}
    \delta(x-\mu) \ast f(x) = f(x - \mu) \,.
\end{equation*}

A Parzen density estimate using a sum of Gaussians each centered at data points $x_i$ is therefore equivalent to a sum of shifted delta impulses convolved with a Gaussian
\begin{equation*}
    \sum_i G_{\sigma} (x - x_i) = G_{\sigma} (x) \ast \sum_j \delta(x - x_i) \,.
\end{equation*}

With respect to applications in binary image processing, this observation is pivotal because every active binary pixel can be understood as a discrete delta impulse located at some coordinate on a regular 2D lattice. In a slight abuse of notation, we may therefore write the density function of a set $\mathcal{X}$ of pixels in (\ref{eq:dens1}) as
\begin{equation*}
    p(x) = \mathcal{X} \ast G_{\xi} \;.
\end{equation*}

Convolution with a Gaussian is a standard and well understood operation in low-level signal processing. In digital image processing, it typically realized using discrete 2D filter masks. In cases where $\xi$ is small, the Gaussian convolution kernel can be separated into two 1D convolution kernels. Successively applying corresponding pre-computed 1D filters of small spatial support will then significantly reduce computation time. In cases where $\xi$ is large, it is more efficient to resort to multiplication in the  Fourier domain or to apply recursive filters \cite{Deriche1992-RIT,vanVliet1998-RGD}. The density $p(x)$ of a set of binary pixels can therefore be computed very efficiently; although it will still require efforts proportional to the number $N$ of pixels, the overall effort per pixel will become very small.

Note that the set of codebook vectors $\mathcal{W}$, too, can be understood as a collection of delta impulses on a discrete lattice. With respect to binary image processing, this is to say that $\mathcal{W}$ can be understood as an image containing $M$ foreground pixels. Accordingly, its density estimate
\begin{equation*}
q(x) = \mathcal{W} \ast G_{\omega}
\end{equation*}
can be computed just as efficiently. On a discrete lattice such as an image pixel array, the two entropy integrals $V(\mathcal{X}; \mathcal{W}) = \int p(x)q(x)\, dx$ and $V(\mathcal{W}) = \int q^2(x)\, dx$ are therefore readily available.

It thus remains to derive a modified update rule for the codebook vectors that assumes the role of (\ref{eq:origupdate}). To this end, we have another look at (\ref{eq:dens2}) and note that
\begin{align}
    \frac{\partial q(x)}{\partial w_k}
    & = \frac{\partial}{\partial w_k} \frac{1}{M} \sum_{j = 1}^M G_{\omega}(x - w_j) \notag \\
    & = \frac{1}{M} \frac{w_k - x}{2 \omega^2} G_{\omega}(x - w_k) \; .
\end{align}

Using this, we differentiate the divergence $D_{cs}(\mathcal{X}, \mathcal{W})$ in (\ref{eq:cauchyschwartz}) with respect to $w_k$, equate to zero, rearrange the resulting terms, and obtain the following update rule:
\begin{align}
    w_k^\text{new} = \;
        & \frac{\int p(x) G_{\omega}(x - w_k) \, x \, dx}{\int p(x) G_{\omega}(x - w_k) \, dx} \nonumber \\
        & - c \cdot \frac{\int q(x) G_{\omega}(x - w_k) \, x \, dx}{\int p(x) G_{\omega}(x - w_k) \, dx} \nonumber \\
        & + c \cdot \frac{\int q(x) G_{\omega}(x - w_k) \, dx}{\int p(x) G_{\omega}(x - w_k) \, dx} \, w_k \label{eq:accupdate}
\end{align}
where $c = \frac{V(\mathcal{X};\mathcal{W})}{V(\mathcal{W})}$.

Note, that in contrast to (\ref{eq:origupdate}), every Gaussian kernel in (\ref{eq:accupdate}) is of the form $G_{\omega}(x - w_k) = \delta(x - w_k) \ast G_{\omega}(x)$. That is, they are now centered at $w_k$, are all of the same variance, and do not specifically depend on the $x_i$ or the other $w_j$ anymore.

On a discrete lattice, this allows for further acceleration. Following common practice in digital signal processing, we can approximate the continuous Gaussian $G_{\omega}(x)$ by means of a pre-computed discrete filter mask $F$ of finite support. Instead of evaluating the integrals in (\ref{eq:accupdate}) over the whole lattice, it then suffices to consider a neighborhood $\mathcal{N}$ of $w_k$ whose size depends on $\omega$. That is, on a discrete lattice, it is valid to assume the approximation
\begin{align*}
   \int p(x) \, G_{\omega}(x - w_k) \, x \, dx  \; & \approx \; \sum_{x \in \mathcal{N}} p(x) \, F(x - w_k) \, x \;
\end{align*}
and similar expressions hold for the other integrals in (\ref{eq:accupdate}).

Concluding our derivation up to this point, we observe that, in contrast to the original update algorithm for ITC, our version for discrete lattices (i) avoids the computation of Euclidean distances and (ii) does not necessitate explicit evaluation of exponentials. On a discrete lattice both these steps can be replaced by highly efficient convolution operations using pre-computed filter masks.

\subsection{ITC for Binary Images}
In this subsection, we discuss implementation details of our accelerated ITC algorithm. To simplify our discussion, we focus once again on binary images and show how they may be processed efficiently.

For binary image processing, our efficient version of the original ITC algorithm draws on an isomorphism between a binary image matrix $\mat{X} = [X_{uv}]$ and a finite set $\mathcal{X}$ of vectors $x_i \in \mathbb{N}^2$. That is, in an implementation of our algorithm as well as in our discussion, we capitalize on the isomorphism
\begin{equation*}
    X_{uv} =
        \begin{cases}
        1, & \text{ if } x_i = [u, v]^T \in \mathcal{X} \\
        0, & \text{ otherwise}
        \end{cases}
\end{equation*}
so that we may use $\mat{X}$ and $\mathcal{X}$ interchangeably. Of course, a similar equivalence also holds for the set of codebook vectors so that we may refer to them using either $\mat{W}$ or $\mathcal{W}$.

\begin{algorithm}[t]
\caption{\label{alg:accict} Efficient ITC for binary image quantization}
\begin{algorithmic}[1]
\STATE randomly initialize $\mat{W}$, for instance by sampling from the foreground pixels of $\mat{X}$
\STATE compute the pdf $p(x)$ using $\mat{P} = \mat{X} \ast G_{\xi}$
\REPEAT
\STATE compute the pdf $q(x)$ using $\mat{Q} = \mat{W} \ast G_{\omega}$
\STATE compute the entropy $V(\mathcal{X}; \mathcal{W}) = \sum_{u,v} P_{uv} Q_{uv}$
\STATE compute the entropy $V(\mathcal{W}) = \sum_{uv} Q_{uv}^2$
\FOR{$k = 1, \ldots, M$}
\STATE compute $w_k^\text{new}$ using (\ref{eq:accupdate})
\ENDFOR
\STATE update $\mathcal{W} = \bigcup_k w_k^\text{new}$
\UNTIL{$D_{cs}(\mathcal{X}; \mathcal{W}) \leq \epsilon$ or $\bigl \lVert \mat{W} - \mat{W}^\text{old} \bigr \rVert \leq \theta$}
\end{algorithmic}
\end{algorithm}

Given these prerequisites, Algorithm~\ref{alg:accict} summarizes how to use ITC in order to cluster binary images into coherent segments. The procedure begins with randomly sampling $M$ codebook vectors $w_j$ from the pixels $x_i$ of the shape image $\mat{X}$. Discrete approximations $\mat{P}$ and $\mat{Q}$ of the two probability density functions $p(x)$ and $q(x)$ are (pre-)computed using convolutions.

Also, since the algorithm is tailored to data on discrete lattices, the integrals in $V(\mathcal{X}; \mathcal{W})$ and $V(\mathcal{W})$ are replaced by finite sums over the elements of $\mat{P}$ and $\mat{Q}$. With these ingredients, the codebook vectors $w_k^\text{new}$ are updated according to equation (\ref{eq:accupdate}). These steps are iterated until the Cauchy-Schwartz divergence between $\mat{P}$ and $\mat{Q}$ falls below a user defined threshold or the updates become very small. In practice, we found that it typically requires less than 20 iterations for the procedure to converges to a stable solution.

\section{Performance Evaluation}

Next, we present and discuss results obtained from experimenting with our accelerated version of information theoretic clustering and the original algorithm according to \cite{LehnSchioler2005-VQU,Rao2009-MSA,Rao2007-ITV}. In addition, we present results obtained from applying $k$-means clustering to the pixels of binary images in order to provide a well known baseline for comparison.

\begin{figure}[t!]
    \begin{center}
    \subfigure[$k$-means]{%
    	\includegraphics[width=0.44\columnwidth]{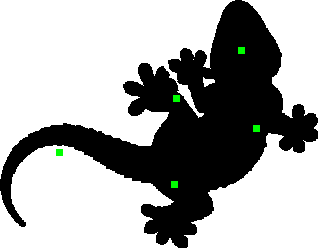}
        \includegraphics[width=0.44\columnwidth]{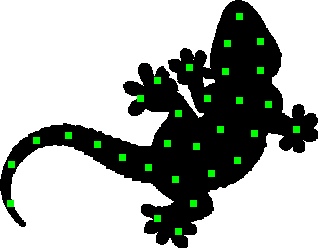}}
    \hfill
    \subfigure[original ITC]{%
    	\includegraphics[width=0.44\columnwidth]{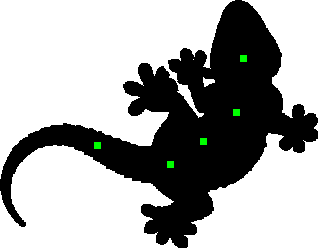}
        \includegraphics[width=0.44\columnwidth]{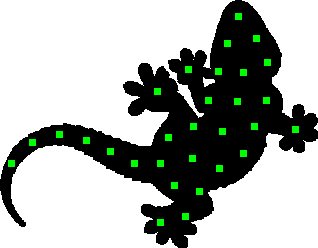}}
    \hfill
    \subfigure[accelerated ITC]{%
    	\includegraphics[width=0.44\columnwidth]{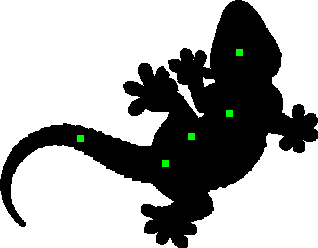}
        \includegraphics[width=0.44\columnwidth]{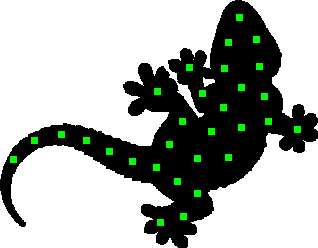}}
    \end{center}
    \caption{\label{fig:shape-exmpls} Examples of the placement of $k = 5$ and $k = 30$ codebook vectors on a binary image of 26.996 foreground pixels. For $k=30$, $k$-means and ITC appear to produce similar results. For $k=5$, however, the mode seeking property of ITC becomes apparent. While $k$-means produces an equidistant covering of the shape, ITC places the cluster centers along its principal curve. Also, while $k$-means cluster centers do not necessarily reside on the shape, those resulting from ITC always do. Finally, the slight differences between the two ITC approaches are due to our use of Gaussian filters of rather small spatial support in the accelerated version. Larger filter masks would increase precision at the cost of increased runtime.}
\end{figure}

Figure~\ref{fig:shape-exmpls} provides an example of the placement of $5$ as well as of $30$ codebook vectors on a binary shape. The results shown here are representative for what we found in our experiments. The original version of ITC and our adapted algorithm produce almost identical results. This was to be expected, since mathematically the two are equivalent. The minor differences visible in the figure can be attributed to the fact, that the original ITC algorithm performs a continuous optimization, while our version places the codebook vectors on a discrete grid of coordinates using filter masks of finite support.

\begin{figure}[t!]
    \begin{center}
    \subfigure[$k$-means]{%
    	\includegraphics[width=0.44\columnwidth]{lizzard-c5-kMeans.png}
        \includegraphics[width=0.44\columnwidth]{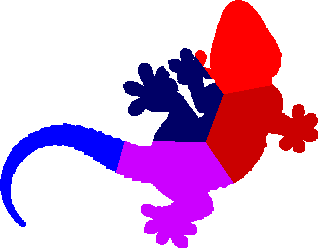}}
    \subfigure[accelerated ITC]{%
    	\includegraphics[width=0.44\columnwidth]{lizzard-c5-aITVQ.png}
        \includegraphics[width=0.44\columnwidth]{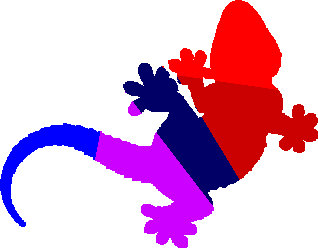}}
    \end{center}
    \caption{\label{fig:cluster-exmpls}. Examples of clusters or shape segments that result from computing $k=5$ codebook vectors using either $k$-means or accelerated ITC.}
\end{figure}

Interestingly, for a larger number of clusters the codebook obtained from $k$-means clustering is rather similar to the ones obtained from ITC. However, $k$-means clustering considers variance minimization for optimization and places cluster centers at local means. It therefore tends to produce blob-like clusters of comparable sizes with equidistant centers. This becomes especially apparent, if a smaller number of clusters are to be determined. In such cases, ITC yields results that are noticeably different.
This difference is further stressed in Fig.~\ref{fig:cluster-exmpls} which, in addition to the the different cluster centers produced by $k$-means and ITC, also shows the corresponding segmentations of the shape.

\begin{figure*}[t!]
    \begin{center}
    \includegraphics[angle=-90,width=0.48\textwidth]{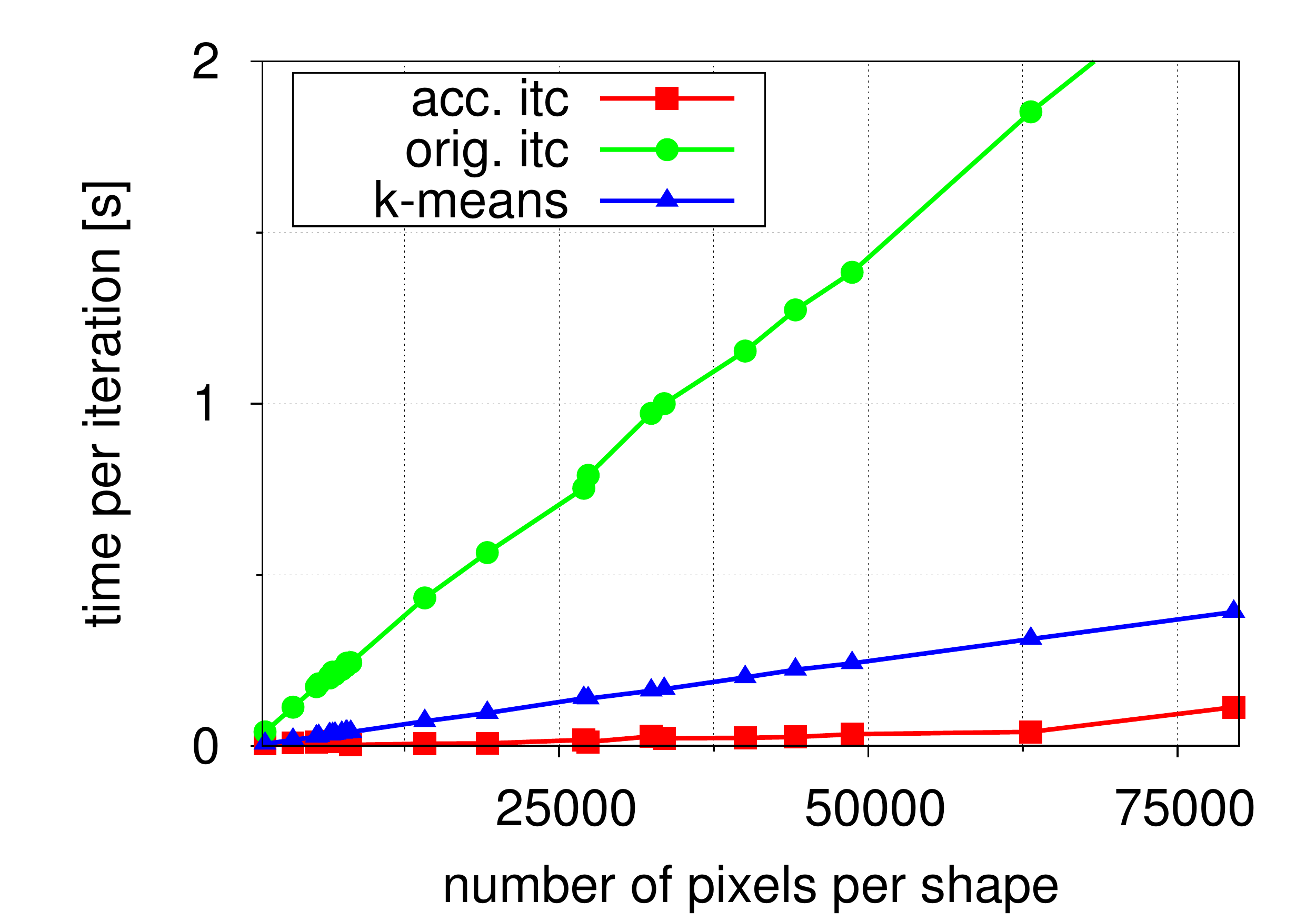}
	\hfill
    \includegraphics[angle=-90,width=0.48\textwidth]{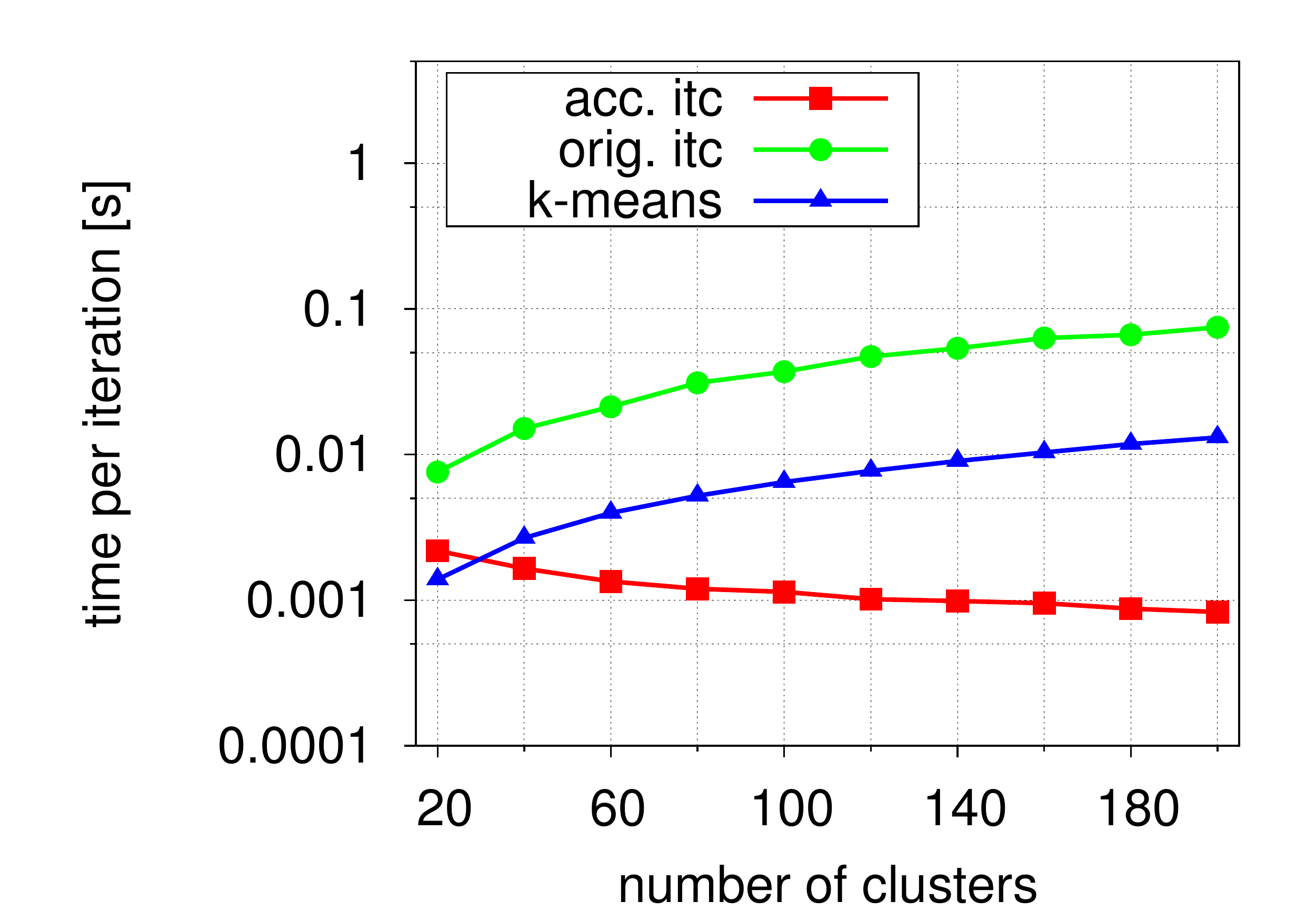}
    \end{center}
    \caption{\label{fig:results}%
     Left panel: average computing time per iteration w.r.t.~shape size. The plots indicate how accelerated ITC, original ITC, and $k$-means perform in extracting 100 codebook vectors from shapes of growing size. %
     Right panel: average computing time per iteration w.r.t.~number of codebook vectors extracted from a shape of about 10.000 pixels. Original ITC and k-means show linear growth (note the logarithmic scale); for accelerated ITC, our way of choosing local support regions (see text) even decreases runtime.}
\end{figure*}

Figure~\ref{fig:results} summarizes quantitative results obtained from runtime experiments with the MPEG-7 shape data set. All our experiments on runtime behavior of the different algorithms were carried out on an Intel Core i3 processor. The variance parameter $\omega$ in our accelerated ITC algorithm was set dynamically, taking into account the number $N$ of pixels in a shape and the number $M$ of codebook vectors to be produced. By choosing $\omega = \sqrt{N/M} / 2$, a smaller codebook will lead to larger discrete filter masks applied in computing ITC according to (\ref{eq:accupdate}). Note that in order to benefit from the idea of using localized filter masks of finite support, it may in fact be preferable to consider convolution masks of smaller diameter. The parameter $\xi$ was set to $\omega / 2$. Finally, all the results we present here were obtained form averaging over 10 trials with different random initializations of $\mathcal{W}$.

Both, the left and right panel of Figures~\ref{fig:results} underline the runtime improvement of accelerated ITC over the original algorithm. For larger choices of $N$ and $M$, the time per iteration of our accelerated algorithm is two orders of magnitude smaller than the time required by the original version. The declining time per iteration of accelerated ITC for a growing number of clusters is a consequence of the above choice of $\omega$ and its impact on the size of  filter masks.

\section{Weighted ITC}

In addition to its favorable runtime characteristics, our adapted ITC algorithm also easily extends towards efficient weighted clustering. Our contribution in this section is to replace the simple Parzen estimate of the density $p(x)$ of the data $\mathcal{X}$ by a more flexible Gaussian mixture model. As we shall see next, for data on a discrete lattice this does not cause any noticeable overhead for our algorithm.

If we extend the model for $p(x)$ in equation (\ref{eq:dens1}) to the more general form
\begin{equation*}
    p(x) = \sum_{i = 1}^N h(x_i) \, G_{\xi}(x - x_i) \label{eq:gmm} \; ,
\end{equation*}
we immediately recognize yet another convolution
\begin{equation*}
    p(x) = \bigl ( h \ast G_{\xi} \bigr) (x) \; .
\end{equation*}

We also note that in our adapted version of the ITC algorithm there is only one place where $p(x)$ is computed explicitly. This happens in the second preparatory step just outside of the loop that iteratively updates the codebook vectors.

With respect to the quantization of binary images, we can therefore introduce a weighting scheme for the pixels in $\mat{X}$ at almost no costs. For instance, we may apply efficient distance transforms \cite{Borgefors1984-AIV} to the foreground pixels of $\mat{X}$ to assign higher weights to the pixels in the interiors of a shape. If we denote this transform by $T(\mat{X})$, the second step in our algorithm simply becomes $\mat{P} = T(\mat{X}) \ast G_{\xi}$.

\begin{figure*}[t]
	\begin{center}
    \subfigure[binary image $\mat{X}$]{%
        \includegraphics[width=0.24\textwidth]{lizzard.png}} \hfill
    \subfigure[chamfer transform $T(\mat{X})$]{%
        \includegraphics[width=0.24\textwidth]{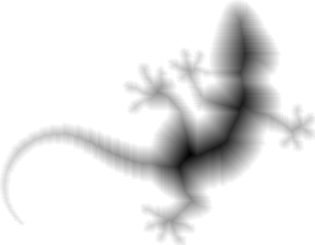}} \hfill
    \subfigure[density $\mat{P} = T(\mat{X}) \ast G_{\xi}$]{%
        \includegraphics[width=0.24\textwidth]{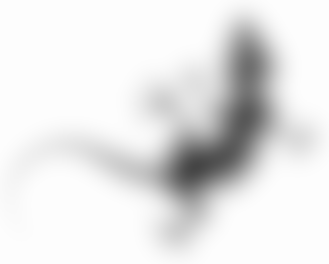}} \hfill
    \subfigure[30 cluster centers]{%
        \includegraphics[width=0.24\textwidth]{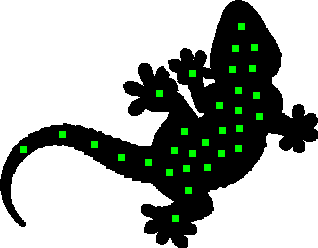}}
    \end{center}
  \caption{\label{fig:weighted-shape-exmpls}%
    Example of weighted, accelerated information theoretic clustering of the pixels of a binary image: (a) the original binary image $\mat{X}$;  (b) the chamfer transform $T(\mat{X})$ yields a gray value image where a pixel's shade of gray indicates its distance to the shape boundary; (c) the correspondingly weighted pdf $\mat{P} = T(\mat{X}) \ast G_{\xi}$ of $\mat{X}$; (d) $k=30$ codebook vectors resulting from weighted ITC; using a distance transform for weighting, causes the cluster centers to move closer to local principal axes of the shape than in the unweighted case in Fig.~\ref{fig:shape-exmpls}.}
\end{figure*}

\begin{figure}[t!]
    \begin{center}
    	\includegraphics[width=0.44\columnwidth]{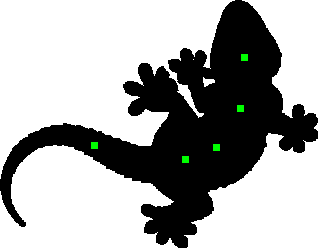}
        \includegraphics[width=0.44\columnwidth]{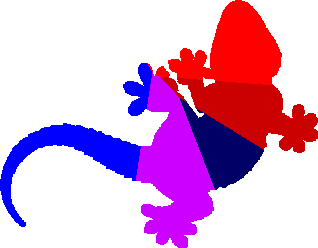}
    \end{center}
    \caption{\label{fig:weighted-cluster-exmpls}. Examples of clusters or shape segments after computing $k=5$ codebook vectors using weighted ITC based on the distance transform illustrated in Fig.~\ref{fig:weighted-shape-exmpls}.}
\end{figure}

Figures~\ref{fig:weighted-shape-exmpls} and \ref{fig:weighted-cluster-exmpls} show examples of the effect of assigning higher weights to pixels in the interior of a shape. In contrast to the normal variant of ITC, the weighted version causes the cluster centers to be concentrated closer to the shape interiors. Since the weighting scheme only impacts the mode seeking component of the algorithm but not the term that causes cluster centers to repel each other, the resulting codebook vectors trace local principal curves of the shape considered here.

\section{Summary and Outlook}

This paper introduced an efficient algorithm for clustering data that reside on a discrete lattice. We demonstrated how costly operations in information theoretic clustering can be be replaced by convolutions. Since convolutions on low dimensional lattices can be computed efficiently, our modified algorithm provides a powerful tool for the quantization of image pixels or voxels.

We illustrated the behavior of the algorithm by means of binary shape images. In experiments carried out to assess the runtime behavior of the modified procedure, we found it to be two orders of magnitude faster than the original algorithm but nevertheless to yield results of similar quality.

Finally, we demonstrated that for data on a lattice, it is straightforward to introduce weighting schemes into the adapted algorithm. Again, given the example of binary images, we showed the effect of applying distance transforms for the purpose of weighting. With this adaptation, the cluster centers produced by the algorithm trace local principal curves or latent features of a shape.

Binary images are a straightforward application domain for our algorithm because their pixels form a distribution of 2D data points. Gray value- and color images, too, can be understood as distributions on low-dimensional lattices. Yet, in practice it would introduce considerable memory overhead to represent them this way. At first sight, the practical applicability of the method proposed in this paper may therefore seem limited. However, this is not the case! On the contrary, modern imaging technology such as tomography or hyper-spectral imaging produces data that meet the prerequisites for our method.

Hyper-spectral images record a whole spectrum of wavelengths for every spatial coordinate of an image. They can thus be thought of as three-dimensional lattices where the data at each grid point is physically related to its neighbors. In this sense, a hyper-spectral image forms a discrete three-dimensional distribution of measurements and accelerated ITC provides a useful tool to uncover latent structures in these data.

In fact, due to their enormous data volumes, hyper-spectral images pose considerable challenges for statistical analysis as traditional methods do simply not apply anymore \cite{Kersting2012-PPO,Thurau2012-DMF}. Yet, there is a lot to gain. For example, in areas such as biology or phytomedicine, hyper-spectral data has been shown to provide new insights into physiological processes within plants \cite{Ballvora2012-DPO,Bauckhage2013-DMA,Bauckhage2012-ATM} and using accelerated ICT promises further interesting results in this demanding big data domain.

Another interesting, though lesser known application area, lies in the field of game AI \cite{Bauckhage2006-WAI,Bauckhage2004-TAF,Sifa2013-AM}. Many modern computer games that are set in simulated 3D worlds require the player to act according to his or her current location. For the programming of convincing artificial game agents this poses the problem of equipping them with a sense of where they are. However, due to the growing size and complexity of virtual, manual annotations that capture all essential areas and aspects may soon become infeasible. Accelerated ICT therefore provides a data-driven approach to the semantic clustering of in-game navigation meshes and first results already underline that it indeed uncovers key locations of in-game maps that make sense from a human player's point of view.

It is in areas like these,  where the method introduced in this paper is of considerable practical benefit and where bridges between advanced machine learning and highly efficient signal processing are in dire need.

\balance

\bibliographystyle{ieee}
\bibliography{literature}

\begin{thebibliography}{10}\itemsep=-1pt

\bibitem{Ballvora2012-DPO}
A.~Ballvora, C.~R{\"o}mer, M.~Wahabzada, U.~Rascher, C.~Thurau, C.~Bauckhage,
  K.~Kersting, L.~Pl{\"u}mer, and J.~Leon.
\newblock {Deep Phenotyping of Early Plant Response to Abiotic Stress Using
  Non-invasive Approaches in Barley}.
\newblock In {\em Proc. Int. Barley Genetics Symposium}, 2012.

\bibitem{Bauckhage2013-DMA}
C.~Bauckhage and K.~Kersting.
\newblock {Data Mining and Pattern Recognition in Agriculture}.
\newblock {\em KI -- K\"unstliche Intelligenz}, 27(4):313--324, 2013.

\bibitem{Bauckhage2012-ATM}
C.~Bauckhage, K.~Kersting, and A.~Schmidt.
\newblock {Agriculture's Technological Makeover}.
\newblock {\em IEEE Pervasive Computing}, 11(2):4--7, 2012.

\bibitem{Bauckhage2006-WAI}
C.~Bauckhage, M.~Roth, and V.~Hafner.
\newblock {Where am I? -- On Providing Gamebots with a Sense of Location Using
  Spectral Clustering of Waypoints}.
\newblock In {\em Proc. Optimizing Player Satisfaction in Computer and Physical
  Games}, 2006.

\bibitem{Bauckhage2004-TAF}
C.~Bauckhage and C.~Thurau.
\newblock {Towards a Fair 'n Square Aimbot -- Using Mixtures of Experts to
  Learn Context Aware Weapon Handling}.
\newblock In {\em Proc. GAME-ON}, 2004.

\bibitem{Borgefors1984-AIV}
G.~Borgefors.
\newblock {An Improved Version of the Chamfer Matching Algorithm}.
\newblock In {\em Proc. ICPR}, 1984.

\bibitem{Cheng1995-MSM}
Y.~Cheng.
\newblock {Mean Shift, Mode Seeking, and Clustering}.
\newblock {\em IEEE Trans. PAMI}, 17(8), 1995.

\bibitem{Comaniciu2002-MSA}
D.~Comaniciu and P.~Meer.
\newblock {Mean Shift: A Robust Approach towards Feature Space Analysis}.
\newblock {\em IEEE Trans. PAMI}, 24(5), 2002.

\bibitem{Deriche1992-RIT}
R.~Deriche.
\newblock {Recursively Implementing the Gaussian and its Derivatives}.
\newblock In {\em Proc. ICIP}, 1992.

\bibitem{Kersting2012-PPO}
K.Kersting, Z.~Xu, M.~Wahabzada, C.~Bauckhage, C.~Thurau, C.~R{\"o}mer,
  A.~Ballvora, U.~Rascher, J.~Leon, and L.~Pl{\"u}mer.
\newblock {Pre-symptomatic Prediction of Plant Drought Stress using
  Dirichlet-Aggregation Regression on Hyperspectral Images}.
\newblock In {\em Proc. AAAI}, 2012.

\bibitem{LehnSchioler2005-VQU}
T.~Lehn-Schioler, A.~Hedge, D.~Erdogmus, and J.~Principe.
\newblock {Vector Quantization Using Information Theoretic Concepts}.
\newblock {\em Natural Computing}, 4(1), 2005.

\bibitem{Rao2009-MSA}
S.~Rao, A.~{de Medeiros Martins}, and J.~Principe.
\newblock {Mean Shift: An Information Theoretic Perspective}.
\newblock {\em Pattern Recognition Letters}, 30(3), 2009.

\bibitem{Rao2007-ITV}
S.~Rao, S.~Han, and J.~Principe.
\newblock {Information Theoretic Vector Quantization with Fixed Point Updates}.
\newblock In {\em Proc. IJCNN}, 2007.

\bibitem{Renyi1961-OMO}
A.~Renyi.
\newblock {On Measures of Information and Entropy}.
\newblock In {\em Proc. Berkeley Symp. on Mathematics, Statistics, and
  Probability}, 1961.

\bibitem{Sifa2013-AM}
R.~Sifa and C.~Bauckhage.
\newblock {Archetypical Motion: Supervised Game Behavior Learning with
  Archetypal Analysis}.
\newblock In {\em Proc. CIG}, 2013.

\bibitem{Thurau2012-DMF}
C.~Thurau, K.~Kersting, M.~Wahabzada, and C.~Bauckhage.
\newblock {Descriptive Matrix Factorization for Sustainability: Adopting the
  Principle of Opposites}.
\newblock {\em Data Mining and Knowledge Discovery}, 24(2):325--354, 2012.

\bibitem{vanVliet1998-RGD}
L.~{van Vliet}, I.~Young, and P.~Verbeek.
\newblock {Recursive Gaussian Derivative Filters}.
\newblock In {\em Proc. ICPR}, 1998.

\end{thebibliography}

\end{document}